\pdfoutput=1

\documentclass[11pt]{article}

\usepackage{EACL2023}

\usepackage{times}
\usepackage{latexsym}
\usepackage[T1]{fontenc}
\usepackage{CJKutf8}
\usepackage[english]{babel}
\usepackage{tabularx}

\usepackage[T1]{fontenc}
\usepackage[utf8]{inputenc}

\usepackage{microtype}
\usepackage{amsmath}
\usepackage{makecell}
\usepackage{graphicx}
\usepackage{booktabs}
\usepackage{fixltx2e}
\usepackage{pifont}
\usepackage{amsfonts }

\title{Detecting Contextomized Quotes in News Headlines by \\Contrastive Learning}

\author{
Seonyeong Song\textsuperscript{1}~~~~Hyeonho Song\textsuperscript{2,3}~~~~Kunwoo Park\textsuperscript{1}~~~~Jiyoung Han\textsuperscript{2}~~~~Meeyoung Cha\textsuperscript{3,2}\\
\textsuperscript{1}Soongsil University~~~~\textsuperscript{2}KAIST~~~~\textsuperscript{3}Institute of Basic Science\\
\texttt{kunwoo.park@ssu.ac.kr,jiyoung.han@kaist.ac.kr}
}

\begin{document}
\begin{CJK}{UTF8}{mj}
\maketitle

\begin{abstract}
Quotes are critical for establishing credibility in news articles. A direct quote enclosed in quotation marks has a strong visual appeal and is a sign of a reliable citation. Unfortunately, this journalistic practice is not strictly followed, and a quote in the headline is often ``contextomized." Such a quote uses words out of context in a way that alters the speaker's intention so that there is no semantically matching quote in the body text. We present QuoteCSE, a contrastive learning framework that represents the embedding of news quotes based on domain-driven positive and negative samples to identify such an editorial strategy. The dataset and code are available at \url{https://github.com/ssu-humane/contextomized-quote-contrastive}.
\end{abstract}

\section{Introduction} 

A direct quotation, a verbatim replication of a speaker's words as opposed to offering news reporters' own opinions, manifests news stories' neutrality, factuality, and objectivity~\cite{ZELIZER+1989+369+388}. Quoting others also adds color to the news with authentic expressions and conveniently establishes authority based on the speakers’ reputation~\cite{newsreportingwriting}. Therefore, a direct quotation constitutes an integral element of news reporting~\cite{nylund2003quoting}.  

More studies have found a link between the use of direct quotations and fake news. Content analyses of news stories document evidence such that deceptive (\emph{versus} trustworthy) news articles contain more direct quotations~\cite{dalecki2009news,govaert2020deceptive}. An equally problematic but less studied concern involving direct quotations is \emph{contextomy}, quoting words out of context in a way that alters the speaker's intention. A previous study argued that contextomy is a "common spin tactic" of news reporters promoting their political agenda~\cite[p.~332]{mcglone2005quoted}.

\begin{table*}[t]
    \small
    \centering
    \begin{tabular}{p{3.75cm} | p{9.2cm} | c}
    \hline
    News headline quote & Body-text quotes & Label\\\hline
    \makecell[l]{
    "이대론 그리스처럼 파탄"\\
    (A debt crisis, like Greece, is \\ on the horizon)} &  
    \makecell[l]{"건강할 때 재정을 지키지 못하면 그리스처럼 될 수도 있다"\\(If we do not maintain our fiscal health, we may end up like Greece) \\"강력한 ‘지출 구조조정’을 통해 허투루 쓰이는 예산을 아껴 필요한 \\곳에 투입해야 한다" (Wasted budgets should be reallocated to areas \\ in need through the reconstruction of public expenditure)}  & Contextomized \\ \hline
    \makecell[l]{"불필요한 모임 일절 자제"\\
    (Avoid unnecessary gatherings \\ altogether)} & \makecell[l]{"저도 백신을 맞고 해서 여름에 어디 여행이라도 한번 갈 계획을\\ 했었는데..." (Since being vaccinated, I had planned to travel somewhere \\in the summer, but ...)\\"행사가 일단 다 취소됐고요..." (Events have been canceled...)\\"어떤 행위는 금지하고 어떤 행위는 허용한다는 개념이 아니라 \\불필요한 모임과 약속, 외출을 일제 자제하고…." (It is not a matter of \\prohibiting or permitting specific activities, but of avoiding unnecessary \\ gatherings, appointments, and going out altogether...)}  & Modified \\\hline 
    \end{tabular}
    \caption{Dataset examples in Korean and their English translations}
    \label{tab:quote_main_examples}
\end{table*}

Some news outlets have been notorious for editorializing and sensationalizing their stories with contextomized quotes in news headlines ~\cite{han2013comparative}. The first example in Table~\ref{tab:quote_main_examples} illustrates contextomy. This example has a headline, "A government handing out money ... `A debt crisis, like Greece, is on the horizon'." The quoted sentence rephrased a financial expert saying in the body text, "If we do not maintain our fiscal health, we may end up like Greece." This is far from word-for-word replication. Instead, the headline reduced the expert's normative claim about government spending and fiscal distress to a blurb that blasted the national leadership, which was on the opposite side of the political spectrum. As such, a contextomized quote in a news headline can serve as an editorial slogan, misinforming public opinion.

\begin{figure}[t!]
    \centering
      \includegraphics[width=.65\linewidth]{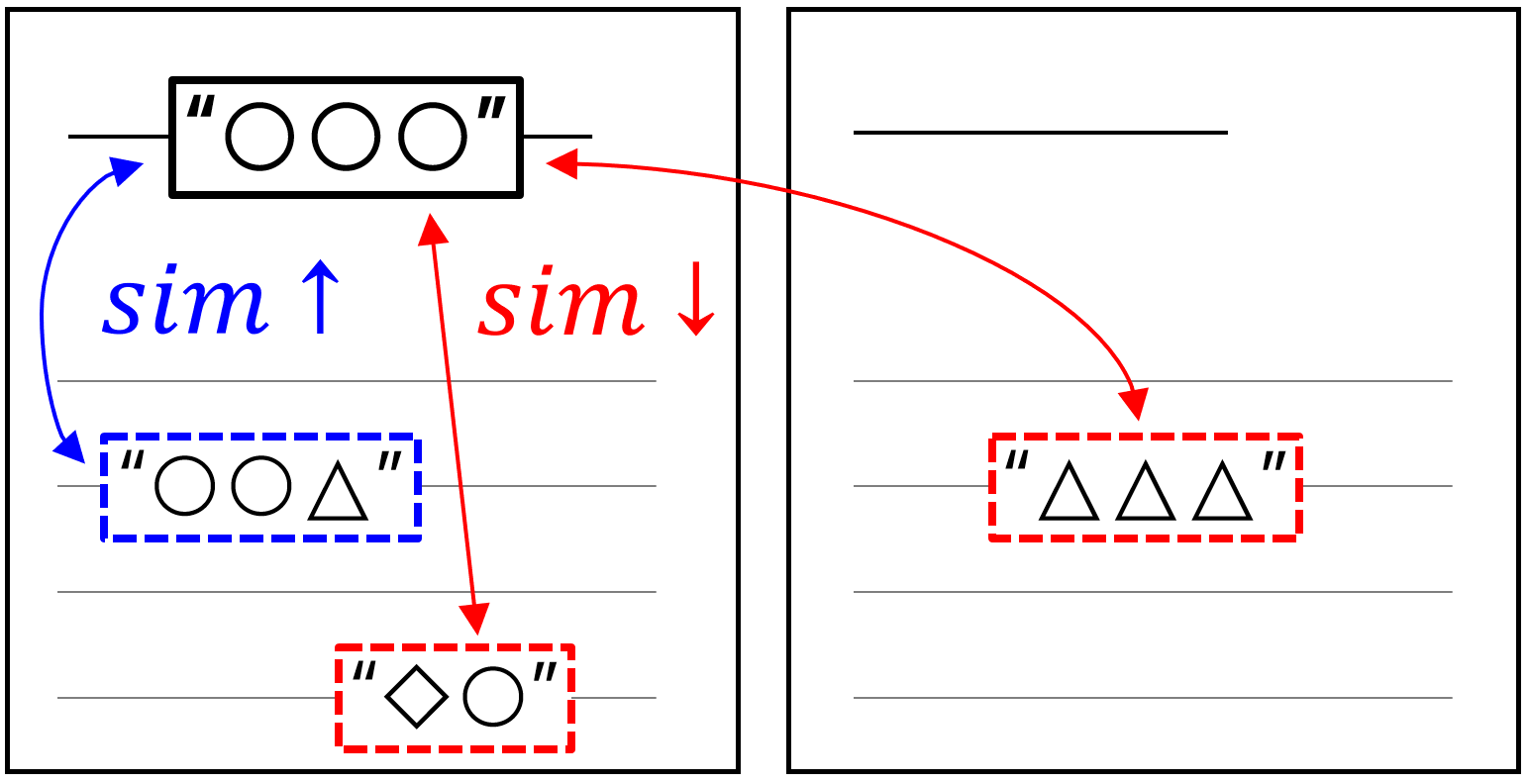}
    \caption{The central idea of QuoteCSE is based on journalism principles, where quotes from news headlines and body text should be matched. The proposed contrastive learning framework maximizes the semantic similarity between the headline quote and the matched quote in the body text while minimizing the similarity for other unmatched quotes in the same or other articles.} 
    \label{fig:intro}
\end{figure}

We propose a new problem of identifying contextomized quotes in news headlines. In contrast to a modified quote, which corrects grammar, replaces unheralded pronouns with proper names, removes unnecessary phrases, and substitutes synonyms, a \textbf{contextomized quote} refers to the excerpt of words with semantic changes from the original statement~\cite{mcglone2005quoted}. 
Hence, the task is to classify whether a headline quote is semantically matched by comparing quotes in the news headline and body text.

To tackle the detection task, we propose using contrastive learning for quote representation, which trains a model to maximize the similarity of samples that are expected to be similar (known as \textit{positive} samples). Simultaneously, the model tries to reduce the similarity between samples that should be dissimilar (aka \textit{negative} samples). Following the recent research in contrastive sentence embedding~\cite{gao-etal-2021-simcse,chuang-etal-2022-diffcse}, we introduce a positive and negative sample selection strategy that is suited to the problem.

Our key idea is illustrated in Figure~\ref{fig:intro}. If a direct quotation appears in a news headline, there should be a quote with the same semantics in the body text. Furthermore, the title quote must be distinct from other quotes in the same article or from quotes in other (randomly chosen) news articles. Since quotes from the same article share common topics, it is more challenging to distinguish a headline quote from those in its body text than to understand semantic differences between quotes from distinct articles. Adopting the `hard' negatives in contrastive loss can help a model learn an effective representation, thereby capturing nuanced semantic differences between quotes. Evaluation experiments show its effectiveness at the target problem as well as its high quality in terms of theoretical measures, such as alignment and uniformity.

Our main contributions are three-fold:
\begin{enumerate}
    \item Based on journalism research and principles, we present a new NLP problem of detecting contextualized quotes in news headlines.
    
    \item We release a dataset for the detection problem based on a guideline constructed by annotators with journalism expertise. The label annotation by three workers achieved Krippendorff's alpha of 0.93.
    
    \item We present QuoteCSE, a contrastive quote embedding framework that is designed based on journalism ethics. A QuoteCSE-based detection model outperformed existing methods, including SimCSE and fine-tuned BERT.
\end{enumerate}

\section{Related Work}

Following the recent success in computer vision~\cite{pmlr-v119-chen20j,He_2020_CVPR,NEURIPS2020_f3ada80d,Chen_2021_CVPR}, previous studies on contrastive sentence embedding focused on how to construct a positive pair by employing data augmentation methods to an anchor sentence~\cite{fang2020cert,giorgi-etal-2021-declutr,2020arXiv201215466W,yan-etal-2021-consert}. 
A recent study showed that a simple dropout augmentation (unlike complex augmentations) with BERT to construct a positive pair could be an effective strategy known as SimCSE~\cite{gao-etal-2021-simcse}. Another study improved the performance by combining SimCSE with masked token detection~\cite{chuang-etal-2022-diffcse}. This study proposes a strategy for selecting positive and negative samples according to journalistic ethics.

\section{Problem and Data}

\paragraph{Research Problem}
Let a given news article be $X:(T, B)$, where $T$ is the news title, and $B$ is the body text. Our task is to predict a binary label $Y$ indicating whether the headline quote in $T$ is either contextomized (1) or modified (0) by referring to the body-text quotes. The detection target is news articles that use at least one direct quotation in the headline and body text.

\paragraph{News Data Collection}
We gathered a nationwide corpus of Korean news articles published through Naver, a popular news aggregator service. Direct quotes in news articles were identified via regular expression.
The dataset contains around 0.4 million news stories published until 2019.

\paragraph{Label Annotation}
Two journalism-major undergraduates were trained to manually label whether a direct quote in the headline is contextomized or modified. The \emph{contextomized} quote refers to the excerpt of words with semantic changes from the original statement. The \emph{modified} quote in a headline keeps the semantics of the original expression but is a different phrase or sentence. A faculty member in mass communication drafted annotation guidelines that stipulated the definitions of contextomized and modified quotations with multiple examples. The annotators reviewed the guidelines and labeled 70 (up to 200) news articles per training session. Inconsistent cases were discussed to reach a consensus. After the eighth iterative training practice over two weeks, the annotators achieved high inter-coder reliability (Krippendorff’s alpha = 0.93 for 200 articles). Then the annotators split the rest and labeled the news articles separately.

We randomly sampled 2,000 news articles for the manual annotation. We ignored cases where the body text includes an identical quote to the one in the headline because its detection can be achieved by a string-matching method without learning.
As a result, the final dataset comprises 814 contextomized and 786 modified samples, leaving a total N of 1,600. Table~\ref{tab:quote_main_examples} presents examples. We investigate contrastive embedding approaches to utilize the 381,206 news articles that remained unlabeled.

\section{Methods}

To predict the label $L$ of $X:(T, B)$, we utilize contrastive embedding and measure the semantic relationship between quotes in the headline and body text. We introduce the main framework.

\subsection{Background: SimCSE}

SimCSE~\cite{gao-etal-2021-simcse} is a contrastive learning method that updates a pretrained bidirectional transformer language model to represent the sentence embedding. Its loss function adapts InfoNCE~\cite{oord2018representation}, which considers identical text with a different dropout mask as a positive sample and the other text within the same batch as negative samples. Formally, the SimCSE loss of $i$-th text $x_i$ is 
\vspace{-1mm}
\begin{equation}
    \label{eq:simcse}
     -\text{log} \frac{e^{\text{sim}(\mathbf{h}_i,\tilde{\mathbf{h}}_i)/\tau}}{\sum^{N}_{j=1}e^{\text{sim}(\mathbf{h}_i,\tilde{\mathbf{h}}_j)/\tau}},
\end{equation}
where $\mathbf{h}_i$ is $x_i$'s embedding\footnote{We applied a 2-layer MLP projection head to the hidden representation corresponding to the [CLS] token in the pretrained BERT.}, $\tilde{\mathbf{h}}_i$ is the embedding of positive sample, $\tau$ is temperature hyperparameter, $N$ is the batch size, and $\text{sim}(\cdot,\cdot)$ is the cosine similarity between embedding vectors.

\subsection{Proposed Method: QuoteCSE}

We propose \textbf{QuoteCSE}, a domain-driven contrastive embedding framework on news quotes. Its contribution is in defining positive and hard negatives according to journalism principles. This framework identifies positive and negative samples for a news headline quote according to the golden rules of journalism: When a direct quotation appears in a news headline, its body text should include a quote that is either identical or semantically similar to the headline quote. The latter form can be a good candidate for contrastive learning, where semantically identical yet lexically different quotes serve as `positive' samples. The other quotes in the body text represent different semantics yet cover the same topic, serving as \textit{hard negative} samples. 

We define the QuoteCSE loss of $i$-th sample $X^{(i)}:(T^{(i)},B^{(i)})$ as
\vspace{-1mm}
\begin{equation}
    \label{eq:quotecse}
-\text{log} \frac{e^{\text{sim}(\mathbf{h}_i,\mathbf{h}^+_i)/\tau}}{\sum^{N}_{j=1}\{e^{\text{sim}(\mathbf{h}_i,\mathbf{h}_j^+)/\tau}+e^{\text{sim}(\mathbf{h}_i,\mathbf{h}_j^-)/\tau}\}},
\end{equation}
where $\mathbf{h}_i$ is embedding of headline quote for $i$-th sample. $\mathbf{h}^+_i$ and $\mathbf{h}^-_i$ are embedding of positive and negative quotes in the same body text $B^{(i)}$. $\mathbf{h}^+_j$ and $\mathbf{h}^-_j$ are embeddings of $X^{(j)}$, other news articles in the same batch ($i\neq j$), which are negative samples.

We applied SentenceBERT (SBERT)~\cite{reimers-gurevych-2019-sentence} to make initial assignments on positive (i.e., semantically identical) and negative (i.e., dissimilar) samples among quotes in the body text. A quote is deemed positive if it appears the most similar to the quote in the news headline. After excluding the positive sample, one quote from the body text was chosen randomly as the negative sample. We removed news articles where the cosine similarity between the anchor and the positive sample is below 0.75 because the news headline quote might be contextomized. Additionally, news articles that did not contain at least two quotes in the body text were eliminated. The remaining 86,275 articles were divided into 69,020, 8,627, and 8,628 for training, validation, and testing of contrastive learning methods.

We compared QuoteCSE with three baseline embedding methods, (i) \textsf{BERT}~\cite{devlin-etal-2019-bert}\footnote{huggingface.co/monologg/kobert}, (ii) \textsf{SBERT}\footnote{huggingface.co/jhgan/ko-sbert-sts}, and (iii) SimCSE. For BERT and SBERT, we used the model checkpoint that was pretrained on a Korean corpus. For SimCSE, we tested two versions. The first version is to train BERT on our news corpus by minimizing Eq.~\ref{eq:simcse} on headline quotes~(\textsf{SimCSE-Quote}). The second version is a publicly available SimCSE embedding pretrained on a corpus on natural language inference in Korean (\textsf{SimCSE-NLI})\footnote{github.com/BM-K/KoSimCSE-SKT}. For QuoteCSE and SimCSE-Quote, we used SBERT for the initial assignments of positive and negative samples. The assignments iteratively get updated for every training step using the target embedding being trained (e.g., QuoteCSE). QuoteCSE and SimCSE-Quote were trained on the 69,020 sizes of the unlabeled corpus with a batch size of 16, which is the upper limit under the computing environment. 

\begin{table}[t]
\centering
\resizebox{.9\linewidth}{!}{%
\begin{tabular}{lcc}
\toprule
                            & F1    & AUC \\ \midrule
BERT                 & 0.665$\pm$0.007  & 0.662$\pm$0.006 \\
SBERT                & 0.44$\pm$0.083     & 0.591$\pm$0.020          \\
SimCSE-Quote                   & 0.69$\pm$0.009  & 0.686$\pm$0.009 \\ 
SimCSE-NLI                       & 0.617$\pm$0.008  & 0.623$\pm$0.008 \\\midrule
BERT fine-tune            &   \textbf{0.754}$\pm$0.006    &\textbf{0.749}$\pm$0.006            \\
QuoteCSE                     &\textbf{0.77}$\pm$0.007 & \textbf{0.768}$\pm$0.008 \\\bottomrule
\end{tabular}
}
\caption{Performance comparison with baselines.}
\label{tab:evaluation}
\end{table}


To assess the role of contrastive learning,
we implemented a binary MLP classifier with a 64-dimensional hidden layer, following an embedding evaluation framework~\cite{conneau-kiela-2018-senteval}. The model takes $\mathbf{u}$, $\mathbf{v}$, $|\mathbf{u}-\mathbf{v}|$, and $\mathbf{u}*\mathbf{v}$ as input, where $\mathbf{u}$ and $\mathbf{v}$ are the embeddings of a news headline quote and the body-text quote most similar to the $\mathbf{u}$, respectively. In deciding $\mathbf{v}$, cosine similarity is used along with the target embedding. The classifier predicts whether the headline quote is contextomized based on a vector relationship between $\mathbf{u}$ and $\mathbf{v}$.

For evaluation, we report the mean F1 and AUC scores by repeating the split process 15 times on the labeled dataset with a ratio of 8:2. As a strong baseline, we also tested a fine-tuned BERT classifier (\textsf{BERT fine-tune}) that takes '[CLS] $q_t$ [SEP] $q_{b,1}, \cdots, q_{b,N_{b}}$ [SEP]' where $q_t$ is the headline quote, $q_{b,i}$ is the $i$-th quote in the body text, and $N_b$ is the number of body-text quotes. Details of the model configuration and computing environment are in Section~\ref{sec:computing}.

\section{Evaluation Results}

Table~\ref{tab:evaluation} presents the evaluation results for the contextomized quote detection. We report the average performance along with standard errors by repeating the experiments using each different random seed. QuoteCSE obtained an F1 of 0.77 and an AUC of 0.76, outperforming the fine-tuned BERT and other contrastive learning methods. Among the baseline models, the fine-tuned BERT model achieved the best F1 of 0.754, which is significantly higher than the performance of the standard contrastive learning methods. The results point to the effectiveness of journalism-driven contrastive quote embedding for the detection problem. 

\begin{table}[t]
\centering
\resizebox{.95\linewidth}{!}{%
\begin{tabular}{cccc}
\toprule
Positive & Hard Negative & F1    & AUC \\ \midrule
QuoteCSE & QuoteCSE &\textbf{0.77}$\pm$0.007 & \textbf{0.768}$\pm$0.008 \\\midrule
SimCSE & QuoteCSE & 0.7$\pm$0.005    & 0.69$\pm$0.004  \\
QuoteCSE & \makecell{$-$}  & 0.674$\pm$0.006   & 0.673$\pm$0.006  \\\bottomrule
\end{tabular}
}
\caption{Ablation results confirm the role of both positive and negative samples in the model.}
\label{tab:ablation}
\end{table}

\paragraph{Ablation experiment}
We examined the importance of positive and negative samples in the QuoteCSE framework by removing each component. The first model is to replace QuoteCSE's positive sample with that of SimCSE, which is an embedding of the anchor text with a different dropout mask. The second model is to ignore the hard negative sample from QuoteCSE. It only differs from SimCSE in the selection of the positive sample. We trained two contrastive embeddings using the 69,020-size unlabeled corpus. Table~\ref{tab:ablation} presents the results. The detection performance of QuoteCSE was reduced significantly by the ablation of the positive and negative samples. The hard negative sample turned out to be more critical to the detection performance, as F1 of the corresponding model decreased by 0.096. The results confirm the necessity of both positive and negative samples in the QuoteCSE framework.

\begin{table}[t]
\centering
\resizebox{.95\linewidth}{!}{%
\begin{tabular}{lccc}
\toprule
           & \makecell{Alignment\\(title-title)}   & \makecell{Alignment\\(title-body)}   & Uniformity   \\ \midrule
BERT     & 0.638 &0.738  &-0.711 \\
SBERT & \textbf{0.227}   &0.329  &-1.356   \\
SimCSE-Quote     & 0.503 & 0.38 &-2.176 \\
SimCSE-NLI &0.319 & \textbf{0.26}  & \textbf{-3.257}      \\ \midrule
QuoteCSE 
& \textbf{0.15} &\textbf{0.194}  &\textbf{-3.562}   \\ \bottomrule
\end{tabular}
}
\caption{Results of alignment (i.e., closeness of positive samples) and uniformity (i.e., even distribution) scores
}
    \label{tab:alignment}
\end{table}

\paragraph{Embedding quality}
We employed two metrics to evaluate the quality of contrastive sentence embeddings~\cite{pmlr-v119-wang20k}. The first is \textit{alignment}, which measures how closely positive pairs are located in the embedding space. The next is \textit{uniformity}, which measures how evenly distributed the target data is. A smaller value denotes a higher embedding quality for both metrics, and their formal definitions are given in Section~\ref{app:alignment_uniformity}. We examined two alignments: (i) between two embeddings from the same headline quote with a different dropout mask (title-title) and (ii) between a headline quote and a positive quote in the body text (title-body).
We measured the three metrics on the test split of unlabeled data. Table~\ref{tab:alignment} shows that QuoteCSE achieves the best result for all types of theoretical measures, implying a high embedding quality.

\paragraph{Error analysis}
\label{app:false-positive}
We identified a common pattern of false positives where a model deems a quote contextomized, which turned out to be modified. They corresponded to instances in which a quote in the headline represents a claim that combines multiple quotes in the body text. For example, in a news article, a headline quote was ``감옥 같은 생활... 음식 엉망 (Prison-like conditions... Poor food)'' which could be referred to multiple quotes in the body text ``삿포로 생활은 감옥처럼 느껴진다 (Living in Sapporo feels like being in prison)'' and ``음식도 엉망이다 (food is poor).'' Since the current detection framework compares a headline quote and another quote in the body text, it could not detect the corner case of a modified quote. Future studies could investigate an approach that considers multiple quotes in the body text.

\section{Conclusion}

Inspired by the importance of direct quotations in news reporting and their widespread misuse, this study proposed a new NLP problem of detecting contextomized news quotes. While there had been studies on quote identification~\cite{Pavllo_Piccardi_West_2018} and speaker attribution~\cite{10.1145/3437963.3441760}, this study is the first to discern a specific type of headline news quote that distorts the speaker's intention and is cut out of context. Not only does it violate journalism ethics~\cite{newsreportingwriting,nylund2003quoting}, but it can also mislead public opinion~\cite{mcglone2005quoted}. Therefore, tackling the problem of detecting contextomized quotes in news headlines can significantly aid the existing efforts to nurture healthy media environments using NLP techniques~\cite{oshikawa-etal-2020-survey}.

Understanding the subtle semantic differences between quotes from news headlines and those from body text is a prerequisite for detecting contextomized news quotes. To assist with this, we introduce QuoteCSE, a contrastive learning framework for quote representation. We specifically tailored SimCSE~\cite{gao-etal-2021-simcse} to the detection of the editorial slogan by proposing a positive and negative sample selection strategy consistent with journalism ethics. In the evaluation experiments, we confirmed the effectiveness of both positive and hard negative samples in the journalism-driven contrastive learning framework. Altogether, the findings imply the crucial role of domain knowledge in tackling computational social science problems.

\section*{Limitations and Future Directions}

First, since this study was done on a monolingual corpus in Korean, the generalizability of the method to other languages is unknown. Future research could replicate this study in other languages to test its broad applicability. Second, the contrastive learning techniques were only tested to a batch size of 16 due to the particular computing environment. To address this limitation, we also tested MoCo-based methods that mitigate the memory limitation~\cite{chen2020mocov2}; however, the results were unsatisfactory (Section~\ref{app:moco}). The effect of large batch sizes might be examined in future studies. Third, there may be corner cases that the current detection framework is unable to handle. Even if a direct quotation in the headline is schematically consistent with a quote in the body text, this by no means guarantees the authenticity of the quoted remark. It could have been made up by the speaker in the first place. Accordingly, future research warrants considering labels on veracity in conjunction with labels on whether they are contextomized or modified. 

\section*{Ethics and Impact Statement}

Despite the limited headline space, journalism textbooks underscore that direct quotations should meet the strict verbatim criterion~\cite{brooks2001art,newsreportingwriting,cappon1982associated}. This verbatim rule renders news stories with direct quotations more credible and factual. The aforementioned instances of contextomized quotes, however, violate this public trust in journalism. We thus propose a new NLP problem of detecting contextomized quotes and aim to better contribute to the development of responsible media ecosystems. This study is an example of how social science theories can be incorporated with NLP techniques. Thus it will have a broader impact on future studies in NLP and computational social science.

We used public news dataset published through a major web portal in South Korea. Our data is considered clean regarding misinformation because the platform implements a strong standard in deciding which news outlets to admit. However, the considered news data is not free from media bias, and the learned embedding may learn such political bias. Therefore, users should be cautious about applying the embedding to problems in a more general context. We have fewer privacy concerns because our study used openly accessible news data following journalistic standards.

{
\section*{Acknowledgement}
K. Park and J. Han are the corresponding authors. This research was supported by the National Research Foundation of Korea (2021R1F1A1062691), the Institute of Information \& Communications Technology Planning \& Evaluation (IITP-2023-RS-2022-00156360, 2019-0-00075: Artificial Intelligence Graduate School Program (KAIST)), and the Institute for Basic Science (IBS-R029-C2). We are grateful to Seung Eon Lee for putting together the dataset and to the reviewers for their detailed comments that helped improve the paper. 
}

\bibliography{eacl2023}
\bibliographystyle{acl_natbib}

\appendix

\section{Appendix}
\label{sec:appendix}

\setcounter{figure}{0}
\setcounter{table}{0}
\counterwithin{figure}{section}
\counterwithin{table}{section}
\renewcommand\thefigure{\thesection\arabic{figure}}
\renewcommand\thetable{\thesection\arabic{table}}

\subsection{Details of model configuration and computing environment}
\label{sec:computing}
We ran experiments on a machine with an Intel(R) Xeon(R) CPU E5-2620 v4 running at 2.10GHz, four TitanXP 12GB GPUs, and 130GB RAM. All models were evaluated on Python 3.9 with the Transformers library (ver. 4.19.4). We ran contrastive learning experiments with the batch size of 16 using Adam with a learning rate of 0.01, and the maximum number of epochs was 10. The parameter size of KoBERT is 92m, and that of the MLP projection head is 87k with a hidden dimension of 100. The temperature of the softmax is 0.05, which is the same as \citet{gao-etal-2021-simcse}. It took 10 and 13 hours to finish SimCSE and QuoteCSE contrastive training, respectively. For the detection task, we trained models with the same configuration. We did not conduct hyperparameter optimization since the dataset is small. Instead, we reported summary statistics of performance by repeating the data split, model training, and evaluation process while varying random seeds (0, 10, 20, 30, 40, 50, 60, 70, 80, 90, 100, 110, 120, 130, 140). 

\subsection{Formal definition of alignment and uniformity}
\label{app:alignment_uniformity}

Alignment is 
\begin{equation}
    \label{eq:alignment}
    \mathbb{E}_{\left ( x, {x}^{+} \right ) \sim {P}_{pos}} \ {{\left\|f(x) - f({x}^{+}) \right\|}^{2}}
\end{equation}
, where $x$ is an anchor text, $x^+$ is positive sample, and $f(\cdot)$ is an embedding function. $P_{pos}$ is the distribution of positive pairs.

Uniformity is
\begin{equation}
    \label{eq:uniformity}
    \log \  \mathbb{E}_{\left ( x,y \right ) \sim {P}_{data}} \   {e}^{-2}{\left\|f(x) - f(y) \right\|}^{2}
\end{equation}
, where ${P}_{data}$ is the distribution of the anchor text.

\subsection{Additional evaluations}

\subsubsection{Momentum-based methods}
\label{app:moco}

\begin{table}[hbt]
\small
\centering
\begin{tabular}{lcc}
\toprule
                            & F1    & AUC \\ \midrule
         \makecell{MoCo: SimCSE}            &0.658$\pm$0.011  &0.667$\pm$0.008 \\ 
 \makecell{MoCo: QuoteCSE}            &0.756$\pm$0.005 & 0.753$\pm$0.006 \\ \bottomrule
    \end{tabular}
    \caption{Momentum-based methods underperform their corresponding general methods.}
    \vspace{-1.5mm}
    \label{tab:appendix:moco}
\end{table}

Our computing environment is limited, such that all models were trained with a batch size of 16.
Since the batch size decides the number of negatives for InfoNCE-based contrastive learning frameworks, it was reported that a larger batch size can result in better performance~\cite{pmlr-v119-chen20j}. To approximate the effects of a larger number of negatives in a batch, we evaluated MoCo-based approaches that keep samples in previous batches as additional negatives with momentum updates~\cite{He_2020_CVPR}. We set the queue size to be 40 according to the observation on the effect of queue size in a previous study~\cite{wu-etal-2022-esimcse}. We make two observations from Table~\ref{tab:appendix:moco} on the evaluation results of contextomized quote detection. QuoteCSE still outperformed SimCSE, but the MoCo versions performed worse than the general version.

\subsubsection{STS benchmark}
\label{app:sts}

To see if the learned embeddings are generalizable, we tested the baseline and proposed models on the KLUE benchmark on sentence similarity~\cite{park2021klue}. Using the same model architecture for the contextomized quote detection, we trained a model to predict a binary label on whether two given sentences are similar.

\begin{table}[hbt]
\small
\centering
\begin{tabular}{lcc}
\toprule
           & F1    & AUC   \\ \midrule
KoBERT     &0.636       &0.659       \\
SimCSE-Quote & 0.633 & 0.662 \\
QuoteCSE   & \textbf{0.775} & \textbf{0.796}  \\ \bottomrule
\end{tabular}
\caption{Evaluation based on the KLUE-STS benchmark indicates the generality of the proposed method.} 
\vspace{-1.5mm}
    \label{tab:klue}
\end{table}

The evaluation results based on the valid dataset are shown in Table~\ref{tab:klue}. QuoteCSE outperforms KoBERT and SimCSE-Quote, suggesting that our model can produce better semantic embedding. 

\subsubsection{Filtering scenarios in the wild}

\begin{figure}[hbt]
    \centering
      \includegraphics[width=.9\linewidth]{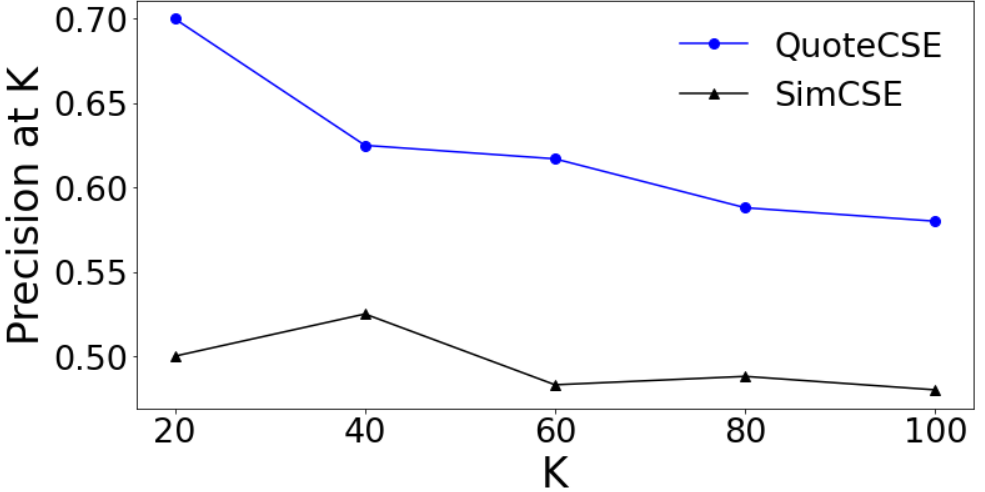}
    \caption{The precision at $k$ results of QuoteCSE and SimCSE suggest QuoteCSE's effectiveness in filtering contextomized quotes in the wild.}
    \vspace{-1.5mm}
    \label{fig:evluation.wild}
\end{figure}

We collected 10,055 news articles published in July and August 2021. To test the proposed model's effectiveness in the wild, we manually evaluated the top-100 news articles regarding the prediction scores of SimCSE-Quote and QuoteCSE, respectively. A high prediction score indicates that a model consider the given news article containing a contextomized quote in headline with a high confidence, therefore this evaluation assumes a scenario of filtering news articles with contextomized quotes.

Figure~\ref{fig:evluation.wild} presents the precision at $k$ of the two models, indicating how many instances turned out to be correct among the top-$k$ examples, which are predicted to be contextomized by a model with a high confidence.
Results indicate that QuoteCSE can achieve a high precision value of 0.7 for the top-20 examples. The precision decreases as its confidence gets lowered, reaching a plateau around 0.6. On the contrary, SinCSE achieved a precision lower than 0.55 even when its confidence is high. The
results suggest the potential of QuoteCSE-based
detection model for filtering contextomized quotes
in the real-world scenario.

\end{CJK}

\end{document}